\documentclass[letterpaper, 10 pt, conference]{ieeeconf}  

\IEEEoverridecommandlockouts                              

\usepackage{times}
\usepackage{graphicx}
\usepackage{amssymb}
\usepackage{url,hyperref}

\title{\LARGE \bf
Road Surface Friction Estimation for Winter Conditions Utilising General Visual Features
}

\author{Risto Ojala$^{1}$ and Eerik Alamikkotervo$^{2}$
\thanks{This work was funded by Henry Ford Foundation Finland.}
\thanks{The authors are with the Department of Mechanical Engineering,
        Aalto University, 00076 Aalto, Finland.}
\thanks{$^{1}${\tt\small risto.j.ojala@aalto.fi}}%
\thanks{$^{2}${\tt\small eerik.alamikkotervo@aalto.fi}}%
}

\begin{document}

\maketitle
\thispagestyle{empty}
\pagestyle{empty}

\begin{abstract}

In below freezing winter conditions, road surface friction can greatly vary based on the mixture of snow, ice, and water on the road.
Friction between the road and vehicle tyres is a critical parameter defining vehicle dynamics, and therefore road surface friction information is essential to acquire for several intelligent transportation applications, such as safe control of automated vehicles or alerting drivers of slippery road conditions.
This paper explores computer vision-based evaluation of road surface friction from roadside cameras.
Previous studies have extensively investigated the application of convolutional neural networks for the task of evaluating the road surface condition from images.
Here, we propose a hybrid deep learning architecture, WCamNet, consisting of a pretrained visual transformer model and convolutional blocks.
The motivation of the architecture is to combine general visual features provided by the transformer model, as well as fine-tuned feature extraction properties of the convolutional blocks.
To benchmark the approach, an extensive dataset was gathered from national Finnish road infrastructure network of roadside cameras and optical road surface friction sensors.
Acquired results highlight that the proposed WCamNet outperforms previous approaches in the task of predicting the road surface friction from the roadside camera images.

\end{abstract}

\section{Introduction}
Winter conditions pose several challenges for operation of road vehicles.
A key challenge is safe manoeuvring in the varying road conditions caused by below freezing winter temperatures.
Studies have shown that winter road conditions have an immense impact on the accident rates of road traffic \cite{wallman2001friction} \cite{strong2010safety}.
Different amounts and mixtures of snow, ice, and water on the road notably affect the road surface friction, i.e. the slipperiness of the road.
Friction between the road and vehicle tyres is a key parameter defining the vehicle dynamics, largely deciding the limits of safe vehicle control and manoeuvring. 
Tyre-road friction is naturally dependent on both road and tyre friction properties, yet the tyre properties are mostly constant, whereas active monitoring is needed for the highly varying road surface friction.
Consequently, evaluation of road surface friction enables different intelligent transportation applications for improving traffic safety.
Such applications can range from adjusting automated vehicle control parameters to alerting drivers of slippery conditions.

Several different approaches have been developed for predicting road surface friction, such as evaluation of vehicle tyre-slip \cite{acosta2017road}, and special optical sensors utilising infrared spectroscopy \cite{jonsson2012infrared}.
In recent years, computer vision methods with regular visible light cameras have gained notable traction.
Cameras are inexpensive sensors, and deep learning approaches have enabled increased accuracy evaluation of the road surface condition.
Development has been aimed at both vehicle on-board cameras, as well as roadside cameras.
Roadside cameras have the benefit of better visibility of the road due to higher installation position, as well as more convenient integration into centralised intelligent transportation data services.

This paper advances roadside camera-based road surface friction prediction by proposing a novel deep learning model, WCamNet, for the task. 
The model consists a hybrid architecture, combining a pretrained visual transformer (ViT), DINOv2, with convolutional blocks.
Previous studies have investigated applying convolutional neural networks (CNNs) as well as the ViT model for task.
Our hybrid architecure feeds general visual features from the pretrained DINOv2 model into a CNN processing pipeline.
This novel approach offers clear accuracy benefits compared to previous state-of-the-art, as indicated by the acquired results.

\section{Related Work}

\subsection{Computer Vision-based Road Surface Condition Monitoring}
Several previous studies have developed methodologies for analysing the road surface condition with camera-based methods, both with on-board cameras and roadside cameras.
Majority of previous studies propose classification methods for the task, providing a general label for the road, e.g. snowy, icy, wet, or dry, which can be used to roughly estimate the road surface friction.
Other methods however aim for higher resolution by directly quantifying a continuous variable as the friction prediction.
Another key characteristic differentiating related works is that several studies are limited to summer conditions, whereas other studies also consider winter conditions and the encompassed highly varying road surface conditions.

Roadside camera-based method for winter conditions was proposed by Jonsson \cite{jonsson2011classification}. 
The proposed method utilised weather information as well as camera images to perform classification of the road surface condition.
Several manually selected features were extracted from images, and principal component analysis was applied to acquire relevant features for the prediction model. 
A simple linear classifier was utilised for prediction.
More recently, Ozcan \textit{et al.} \cite{ozcan2020road} applied deep learning, a VGG16 \cite{simonyan2014very} CNN model for classifying road condition from roadside cameras in winter conditions.
Similarly, Grabowski \textit{et al.} \cite{grabowski2020system} evaluated several CNN models for the same task. 
In their tests, DenseNet201 \cite{huang2017densely} demonstrated the best performance, with VGG19 achieving the second best results.

In summer conditions, with the evaluation limited to dry and wet roads, Sirirattanapol \textit{et al.} \cite{sirirattanapol2019bangkok} have proposed a custom CNN architecture for classifying the road condition from a roadside camera.
Similarly, Abdelraouf \textit{et al.} \cite{abdelraouf2022using} classified dry and wet roads from a roadside camera with deep learning.
However, they applied a ViT model for the task, and they also proposed a novel architecture for fusing inputs from several cameras monitoring the same area.

Development has also been active around computer vision methods aimed for on-board use, utilising a forward facing camera installed on a vehicle.
In the study of Nolte \textit{et al.} \cite{nolte2018assessment}, ResNet50 \cite{he2016deep} and InceptionV3 \cite{szegedy2016rethinking} CNN models were applied to classify road surface condition. 
Winter conditions were included in the analysis.
Wang \textit{et al.} \cite{wang2021road} also developed a model for classifying road surface condition in wintertime. 
They trained a segmentation CNN model with images of different road conditions, and during inference the road condition with most segmented pixels was chosen as the classification.
Vosahlik \textit{et al.} \cite{vosahlik2021self} trained a ResNet50 CNN regression model to predict the vehicle tyre-road friction in winter conditions.
They trained their model based on ground truth values from a slip-based friction estimation method.
Ojala and Seppänen \cite{ojala2024lightweight} proposed a custom CNN regression model, SIWNet, for predicting road surface friction in winter conditions with built-in uncertainty evaluation.
Their model was trained based on ground truths from an optical friction sensor.

On-board methods developed for summer conditions have featured the work of Šabanovič \textit{et al.} \cite{vsabanovivc2020identification}, who developed a custom CNN model capable of classifying the road surface condition.
Similar work has been carried out by Cordes \textit{et al.} \cite{cordes2022roadsaw}, who also published an extensive open dataset called RoadSaw, which featured ground truth values from an optical water film height sensor.
Du \textit{et al.} \cite{du2023pavement} have developed regression-based road surface friction estimation for summer conditions.
They tested several CNN architectures for the task, with VGG19 \cite{simonyan2014very} providing the most accurate results.
They collected ground truth road surface friction information with a commercial contact-based monitoring system.

During winter, road areas typically exhibit different surface condition and friction properties in different sections of an image, due to for example tyre tracks in snow or differences in water film height due to the road elevation profile.
Some studies have addressed this problem by increasing the spatial resolution of the predictions.
To overcome this issue, Roychowdhury \textit{et al.} \cite{roychowdhury2018machine} proposed an on-board approach where they split the image of the road into a grid, and road surface condition in each cell was classified separately.
Pesonen \cite{pesonen2023pixelwise} utilised partially supervised training to develop a deep learning segmentation model to evaluate the road surface friction for each pixel from an on-board camera.
An optical road friction sensor was used to acquire partial labels for the images.
Higher spatial resolution in the prediction is useful in on-board use cases, since the information can be utilised in for example lateral control.
For roadside camera applications however, a single friction value for the entire road area could be sufficient information.
This is due to the probable applications being mostly related to longitudinal control, such as instructing passing vehicles to operate at lower speeds and avoid sharp decelerations.

\subsection{General Visual Features}
With the emerge of larger deep learning models for computer vision, there have been notable efforts in extracting increasingly general visual features. 
This task is motivated by the need for generally applicable computer vision models, which would require minimal further training for adoption to different tasks.
Visual transformer (ViT) models \cite{dosovitskiy2020image} have proven effective for the task.
When trained with large quantities of data in an self-supervised manner, the models are capable of learning rich feature representations for downstream tasks.
These types of models are commonly referred to as foundation models \cite{bommasani2021opportunities}. 

A wide variety of vision foundation models have been published. 
Some are fully self-supervised \cite{zhou2021ibot,he2022masked,zhou2022mugs,assran2022masked,caron2021emerging,oquab2023dinov2} while others use weak supervision from image captions  \cite{radford2021learning,cherti2023reproducible,ilharco_gabriel_2021_5143773}. 
Currently, DINOv2 \cite{oquab2023dinov2} provides the best performance both in image-level downstream tasks like classifications and dense downstream tasks like segmentation.
Due to its capabilities, DINOv2 was chosen as the backbone for downstream friction prediction in this paper. 

The suitability of foundation model features for friction estimation has not been explored in the previous works. 
However, these features are proven to generalise across domains, like different lighting conditions, scales, and environments while encoding high-level information like depth and object class accurately \cite{oquab2023dinov2}. 
Thus the foundation model features are likely suitable also for estimating friction from road appearance in varying lighting and weather conditions. 

\subsection{Research Gap}
Previous works on road surface condition monitoring have developed and extensively evaluated the capabilities of different CNN models, as well as the ViT model.
However, no previous works have explored utilisation of the recent foundation models for the task.
We propose the novel WCamNet model for estimating road surface friction from roadside camera images.
The model combines general visual features acquired with the DINOv2 foundation model, as well as a CNN processing pipeline trained for extracting relevant features for the friction prediction task.
Our results demonstrate that the proposed architecture achieves higher accuracy in the road surface friction prediction task than the state-of-the-art CNN and ViT models or the pretrained DINOv2 foundation model.

\section{Methods}
\subsection{Problem Formulation and Dataset}
This paper focuses on the application of deep learning for analysing the road surface condition from a roadside camera image.
More specifically, the task is formulated as a regression problem, where the deep learning model predicts the road surface friction as a scalar value, with an image as the input.
The aim of this work was to develop a novel prediction model for this task, which outperforms previous state of the art in terms of accuracy.
To facilitate development, a large dataset was gathered from the national Finnish road infrastructure network consisting of roadside cameras and optical road friction sensors \cite{dsc111, dsc211}.
Open access to the data was provided by Fintraffic, the operator of the equipment.

A total of 31 pairs of roadside camera stations and weather stations around Finland were utilised for data collection.
Data samples were gathered at approximately 20 minute intervals for roughly two months in February-March of 2023.
From each station, 1-2 cameras were recorded depending on the number of camera installations in the station.
If 2 cameras were used from the same station, they were looking at opposite directions of the road.
A total of 58 cameras were recorded for the dataset.
Vast majority of the cameras had a resolution of 1280x720 pixels, with some cameras recording at different resolutions.
Highest resolution of the cameras was 1920x1080 pixels, whereas lowest resolution of a recorded camera was 800x450 pixels.
Similarly to the camera stations, each weather station contained 1-2 optical road friction sensors.
If a station contained 2 road friction sensors, their readings were averaged to reach a single ground truth value for the road surface friction.
Out of the 31 weather stations, 17 stations were equipped with a single road friction sensor, whereas 14 stations had two sensors.
The road friction sensors reported friction as a continuous value, \textit{grip factor}, with a measurement range of 0.09-0.82.
Lower values corresponded to slippery conditions, and higher values indicated better tyre grip.
To enable more convenient processing, the readings were scaled to values between 0-1, defined as \textit{friction factor}, $f$.

\begin{figure}
\centering
\includegraphics[width=0.49\textwidth]{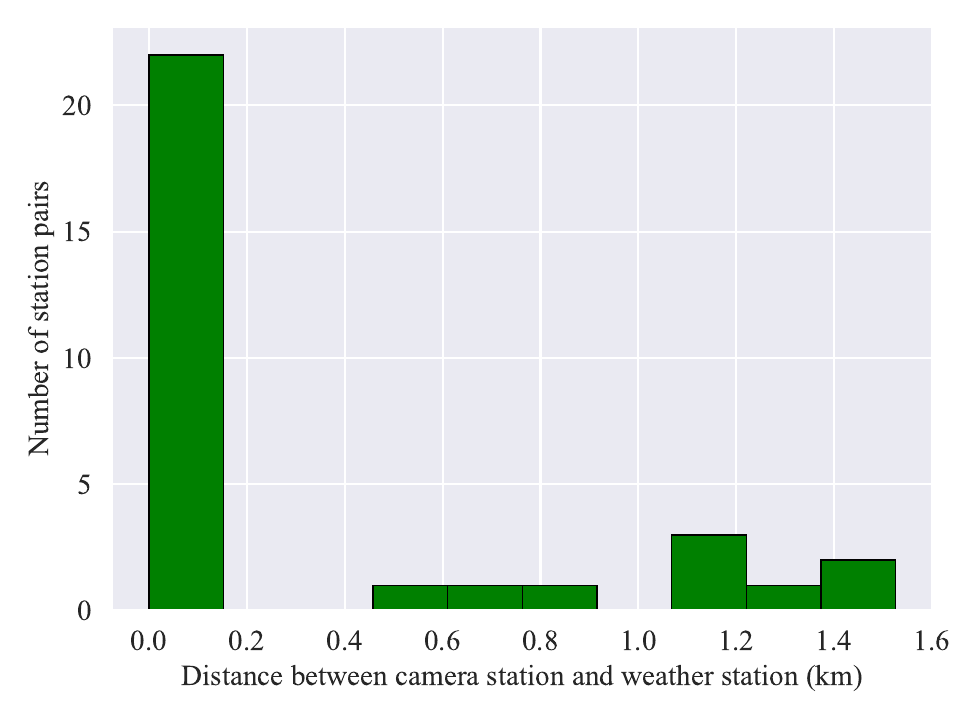} 
\caption{Distribution of distances between the corresponding camera and weather stations.}
\label{fig:distances}
\end{figure}

\begin{figure}
\centering
\includegraphics[width=0.49\textwidth]{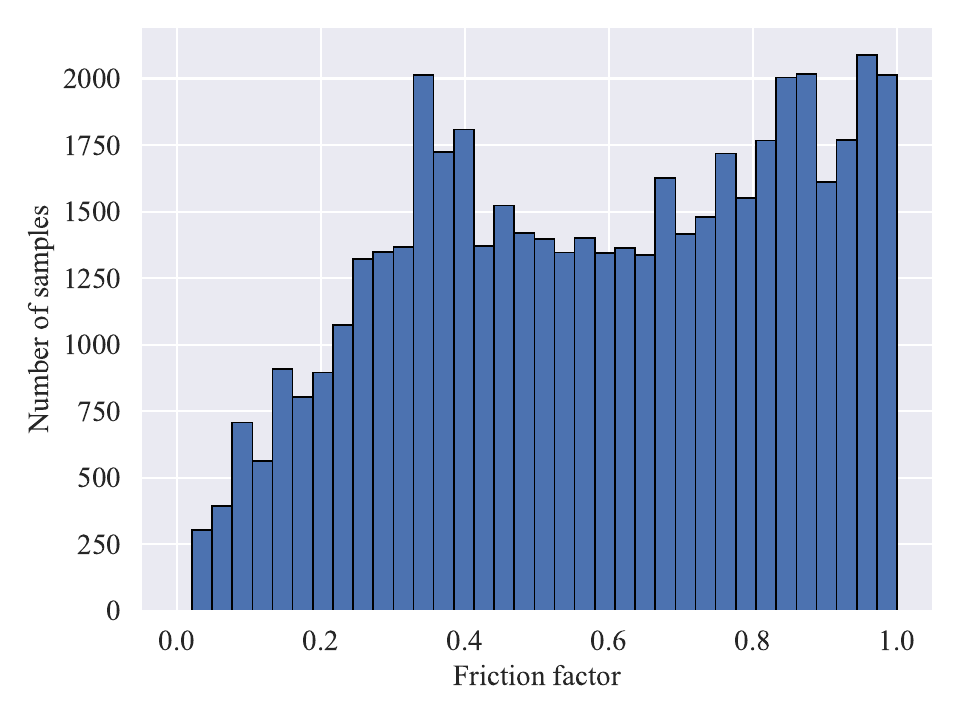} 
\caption{Distribution of friction factor values in the dataset.}
\label{fig:f_distribution}
\end{figure}

Optical road friction sensor information from weather stations was used to label the images captured from the corresponding roadside camera station.
Samples of the gathered data presented in Figure \ref{fig:samples}.
The corresponding weather stations and roadside camera stations were located up to 1.5 km from one another, with majority of corresponding stations in immediate vicinity of one another.
Distribution of distances between the corresponding stations are visualised in Figure \ref{fig:distances}.
The gathered data was initially heavily biased on certain friction factor values, since for example dry asphalt conditions were prevalent across the data collection period. 
Therefore, random weighted sampling was carried out on the gathered data to improve the balance of the dataset.
Sampled from the initially gathered data of over 160 thousand images, the final dataset contained a total of 48791 images with corresponding friction factor labels. 
The distribution of the friction factor values in the final dataset is presented in Figure \ref{fig:f_distribution}.

{\setlength\tabcolsep{2 pt}
\begin{figure*}
\centering
\begin{tabular}{ccc}
 \includegraphics[width=0.33\textwidth]{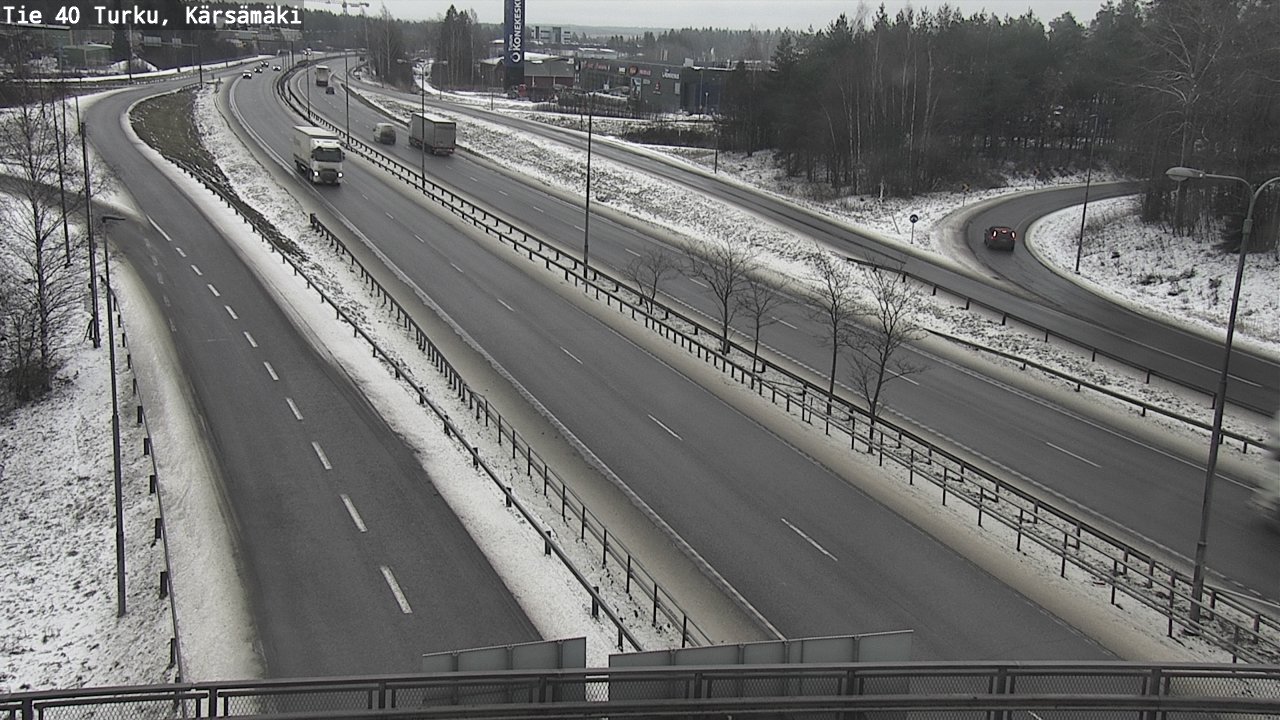} & 
   \includegraphics[width=0.33\textwidth]{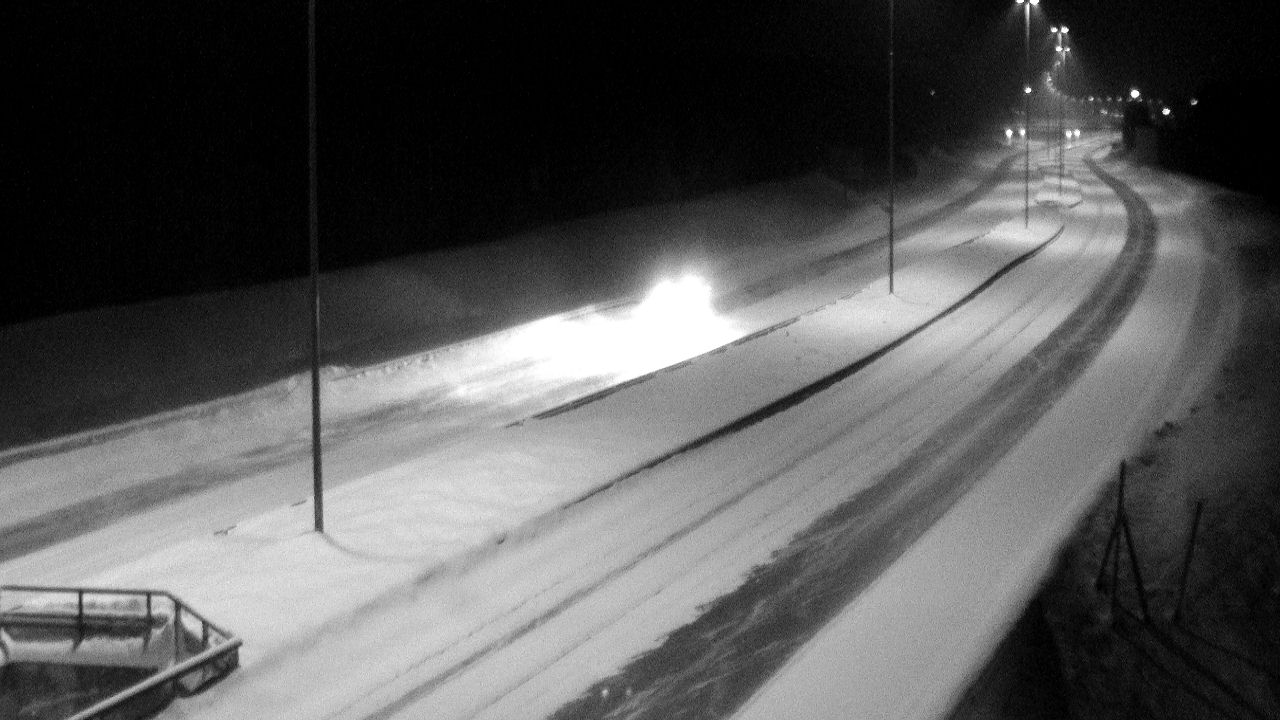} &
    \includegraphics[width=0.33\textwidth]{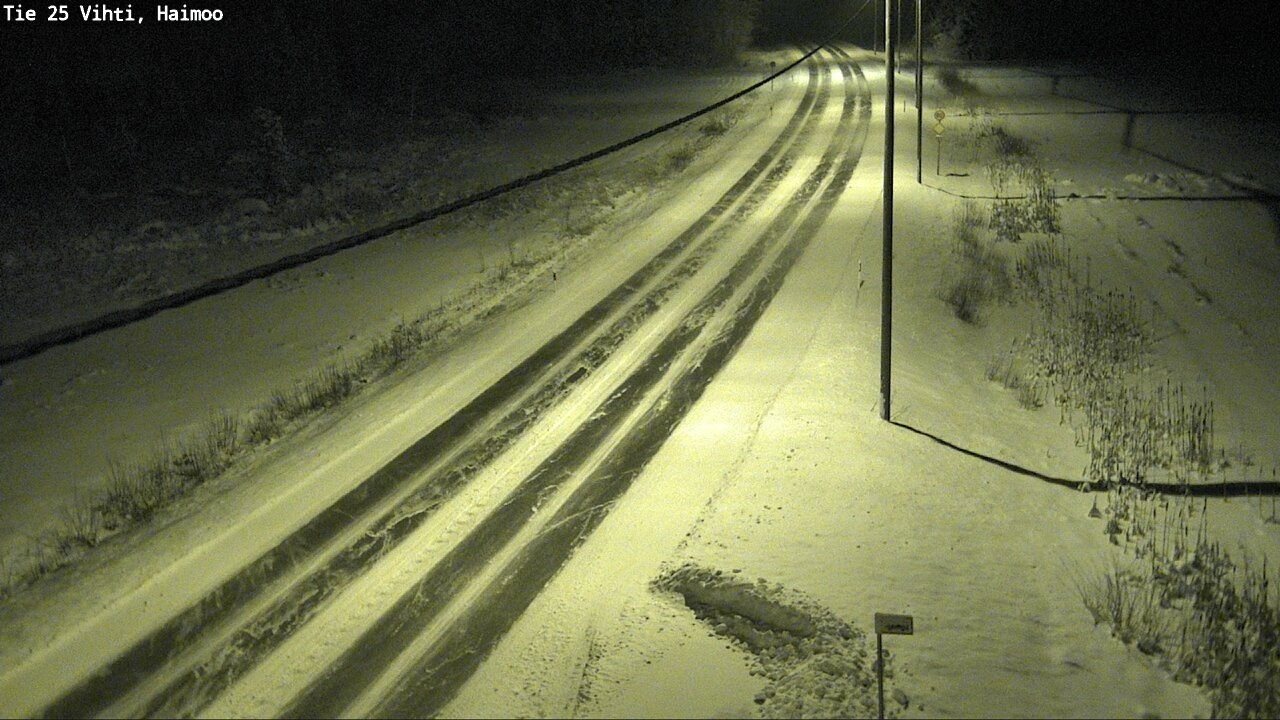} \\
 $f = 0.89$ & $f = 0.32$ & $f = 0.34$ \\
 \includegraphics[width=0.33\textwidth]{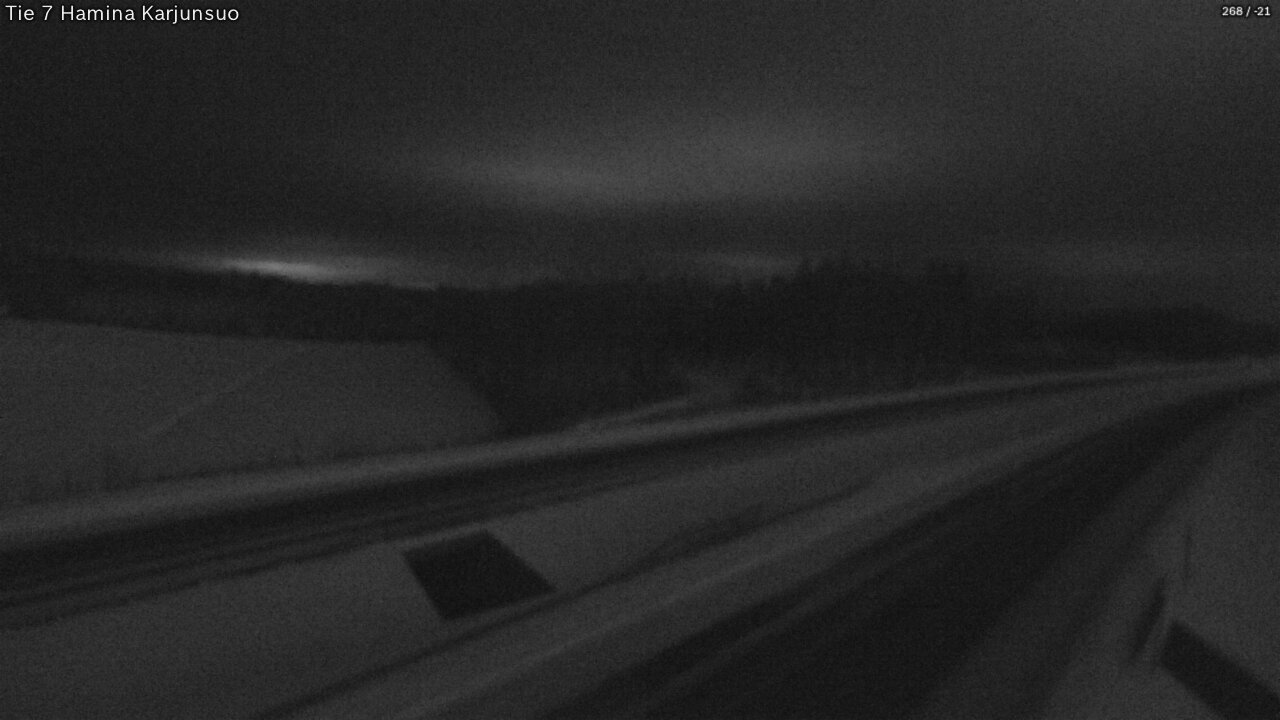} &
  \includegraphics[width=0.33\textwidth]{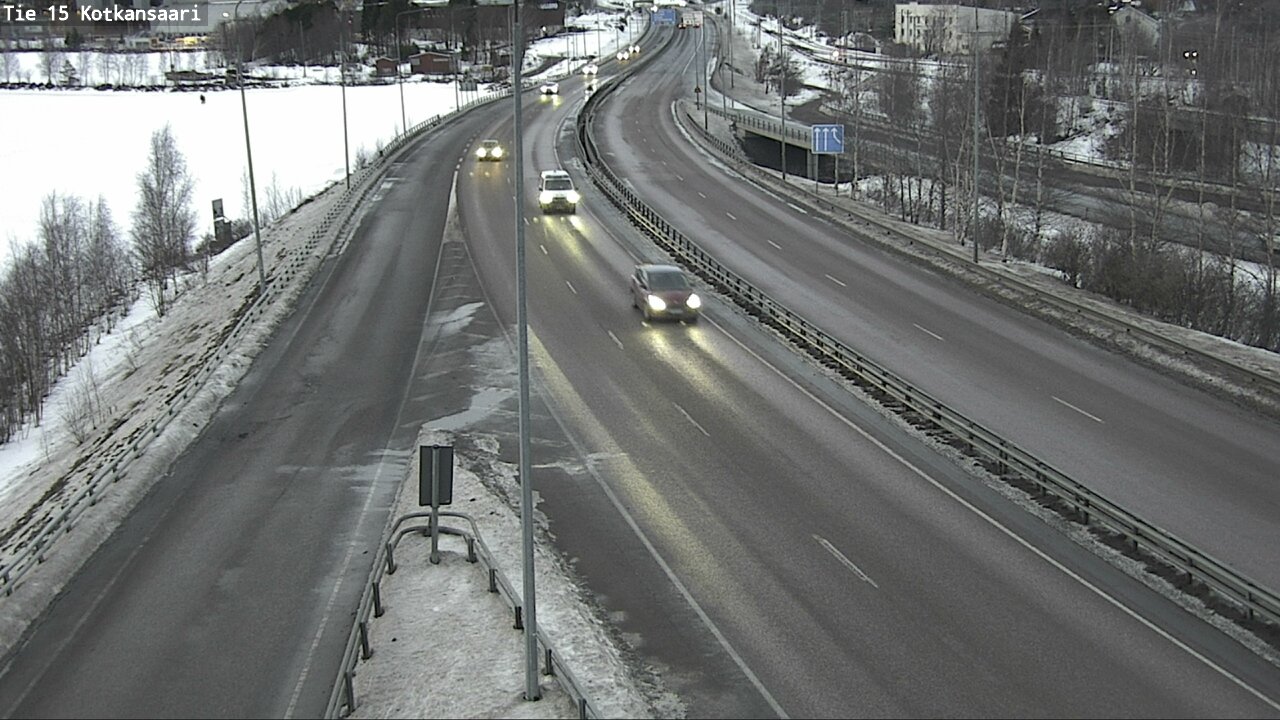} &
   \includegraphics[width=0.33\textwidth]{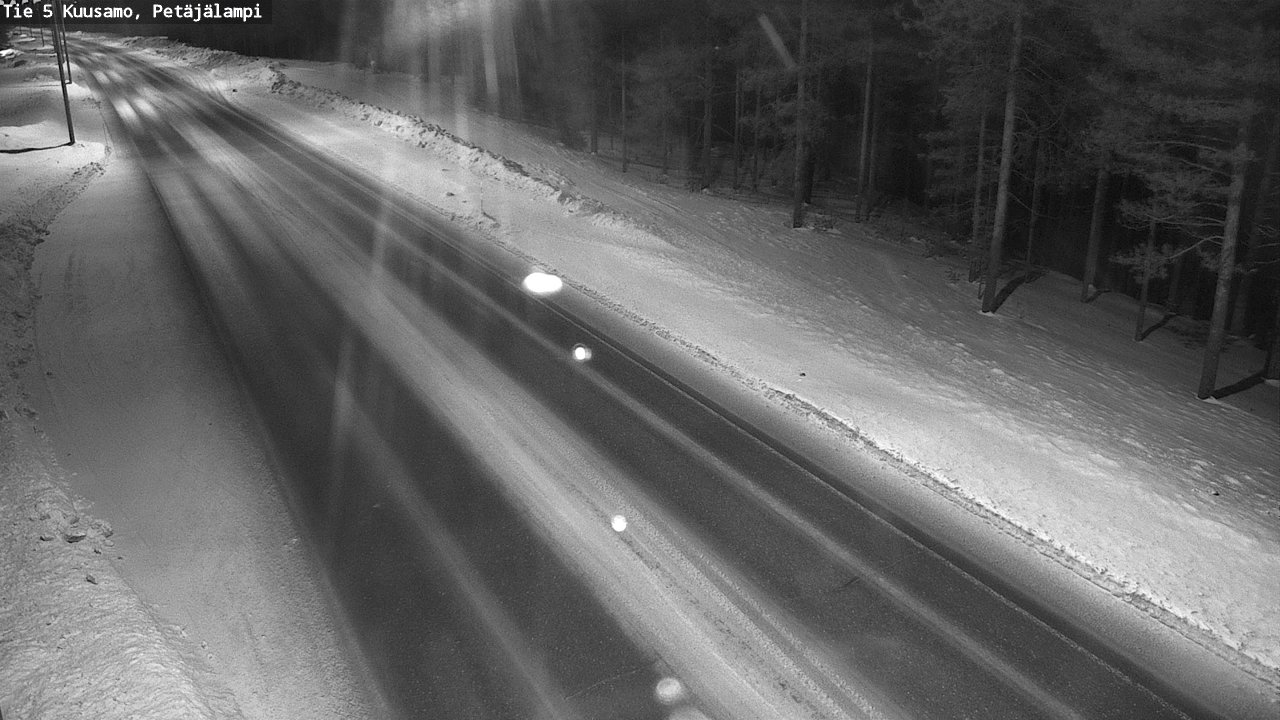} \\
 $f = 0.95$ & $f = 0.95$ & $f = 0.51$ \\
 \includegraphics[width=0.33\textwidth]{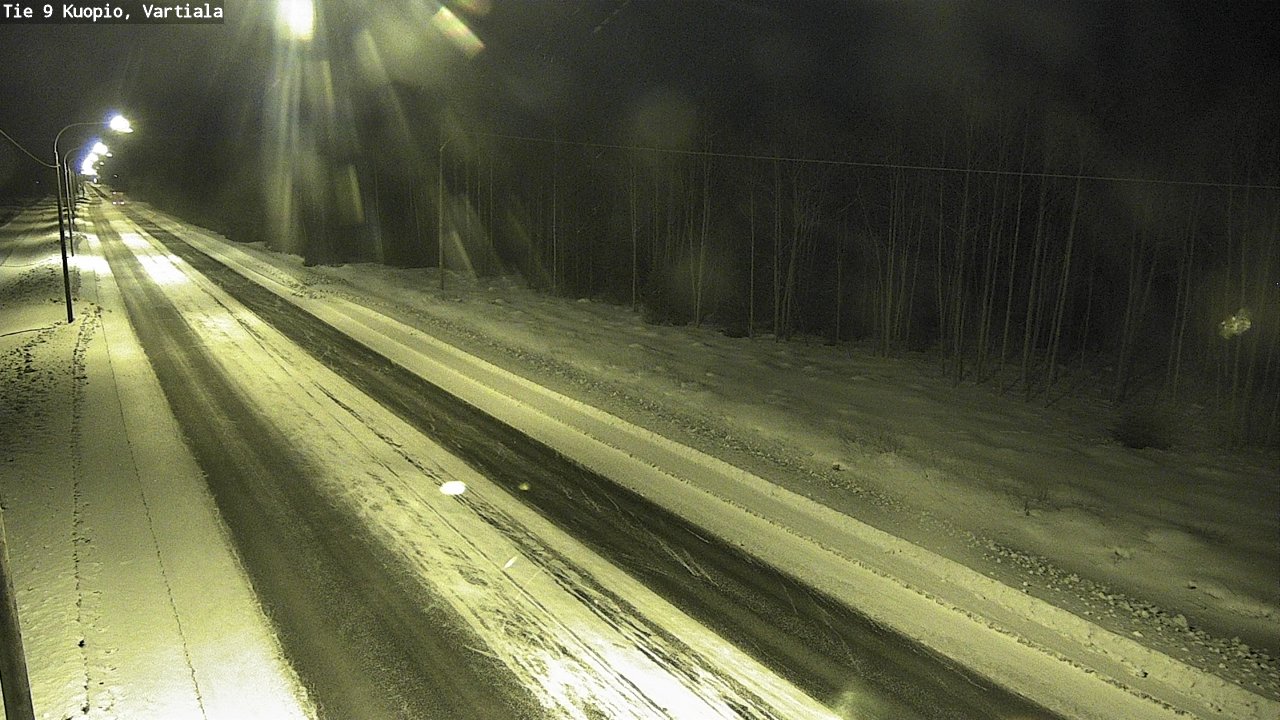} &
  \includegraphics[width=0.33\textwidth]{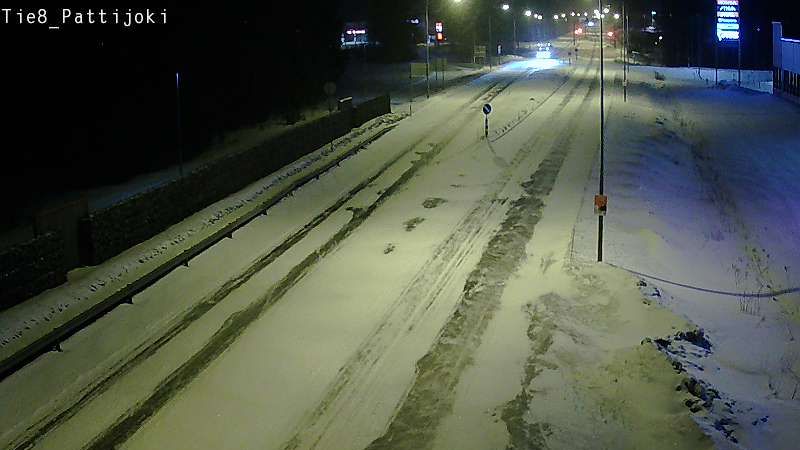} &
   \includegraphics[width=0.33\textwidth]{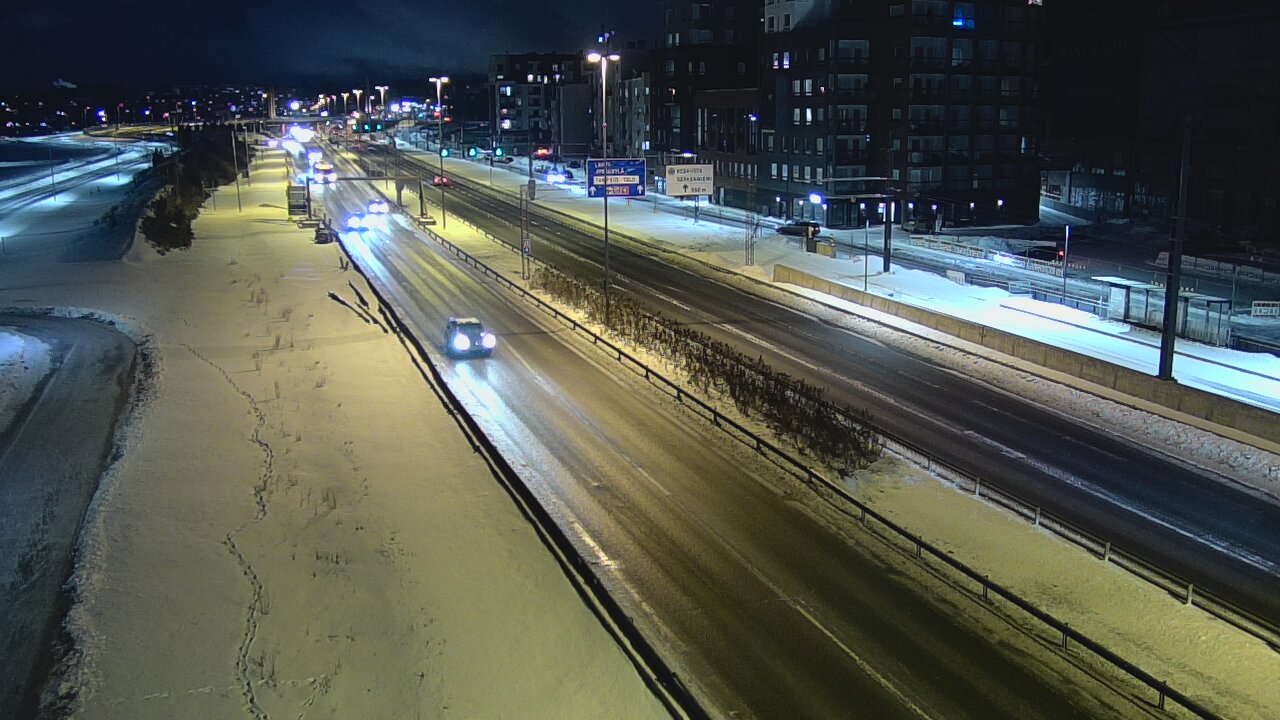} \\
 $f = 0.74$ & $f = 0.45$ & $f = 0.15$ \\
 \includegraphics[width=0.33\textwidth]{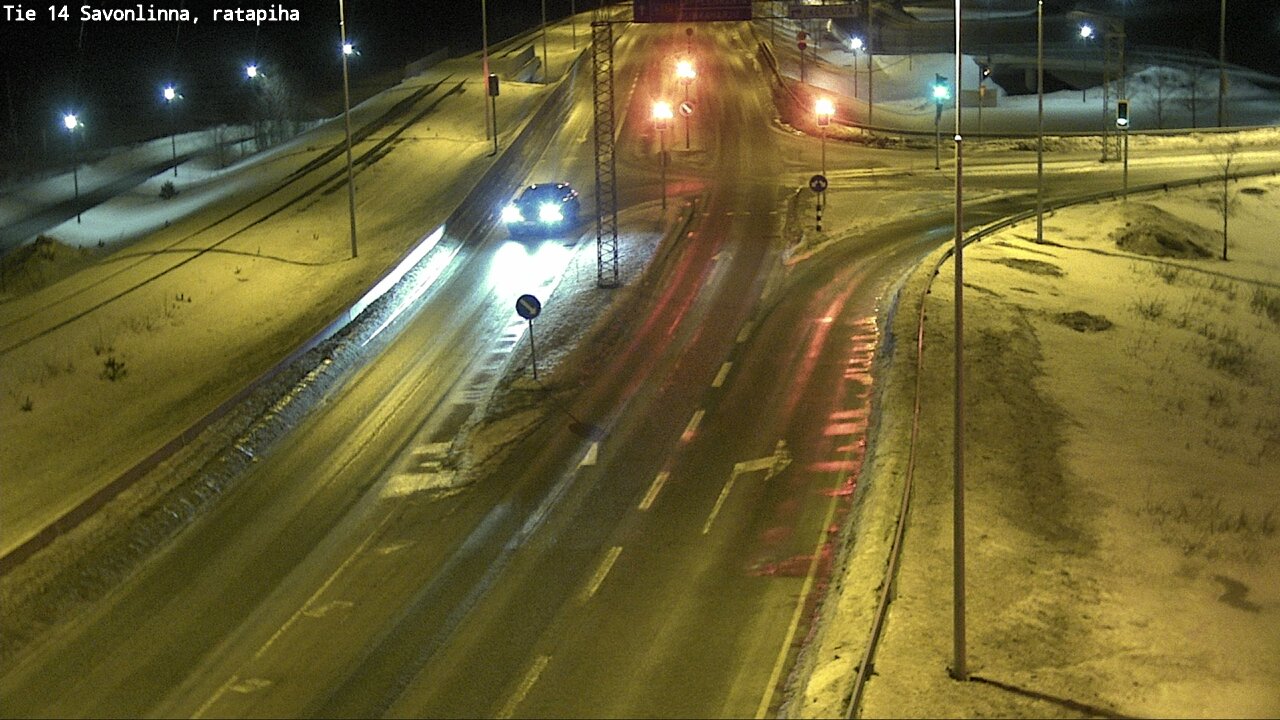} &
  \includegraphics[width=0.33\textwidth]{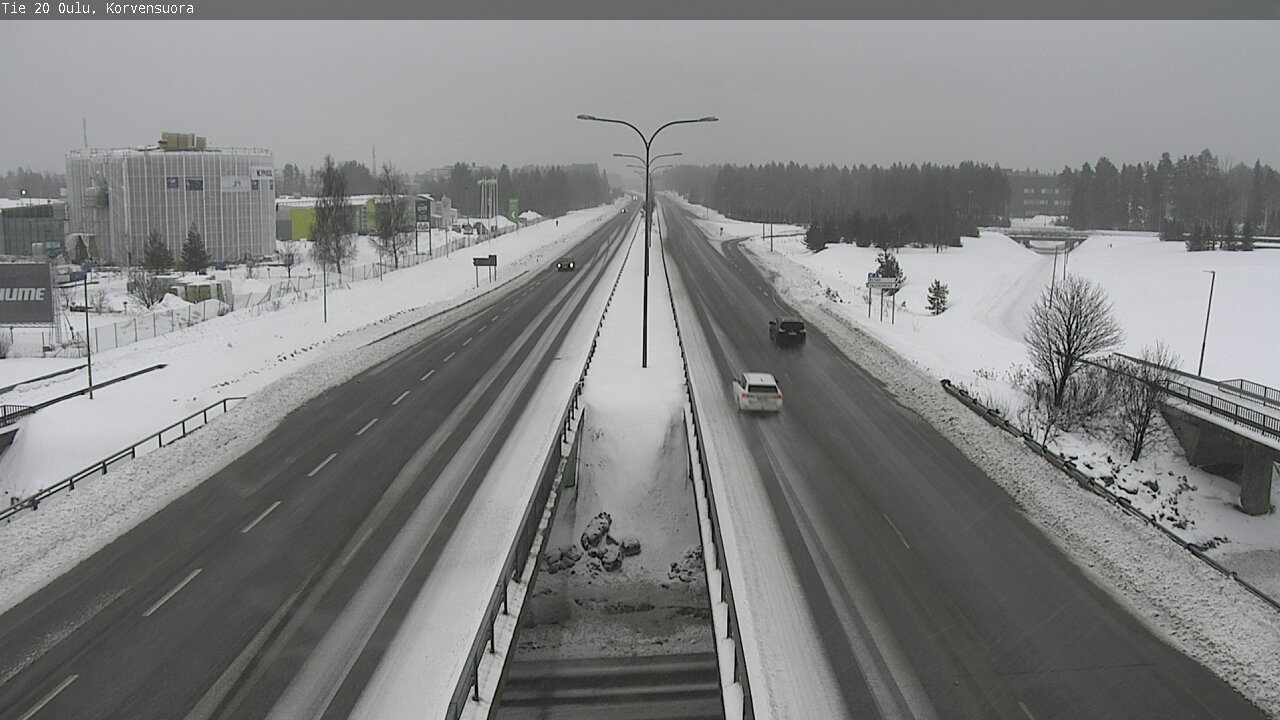} &
   \includegraphics[width=0.33\textwidth]{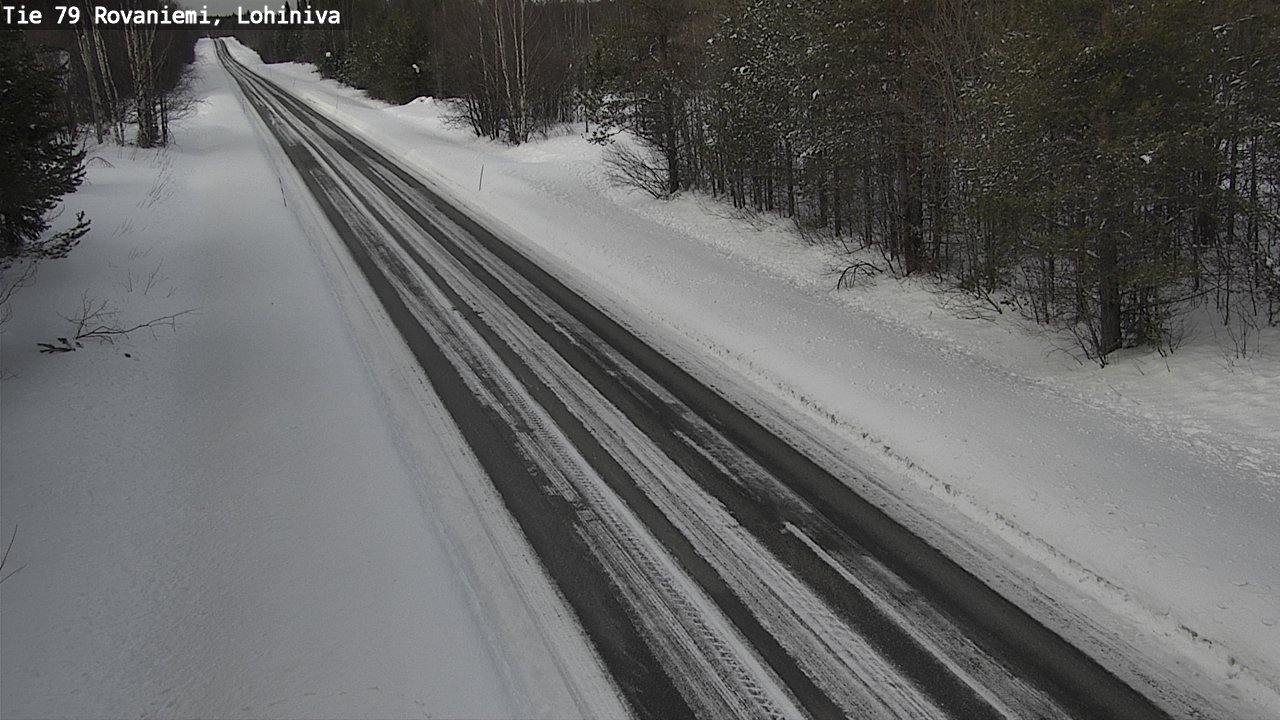} \\
 $f = 0.63$ & $f = 0.77$ & $f = 0.56$ 
 
\end{tabular}
\caption{Samples of image data with corresponding friction factor values.}
\label{fig:samples}
\end{figure*}
}

\subsection{Model Architecture}
Our proposed WCamNet model infuses general visual features from a foundation model with a CNN processing pipeline.
Major motivation for the model was the fact that standalone CNN architectures lack attention mechanisms due to local feature extraction, consequently being unable to properly focus on the relevant sections of the images.
This is detrimental for road surface friction evaluation, since the models are incapable of focusing on the road areas.
Consequently, standalone CNNs are not optimally suited for the prediction task.
WCamNet applies a visual foundation model, DINOv2, to fuse scene understanding and attention for the CNN processing.
Figure \ref{fig:dinov2_features} visualises DINOv2 patch tokens acquired for resized samples from the gathered data, reduced to three dimensions with principal component analysis.
The samples highlight the effectiveness of the foundation model in segmenting varying areas of the scenery, which offers a useful attention mechanism for futher processing.
Furthermore, foundation models have learned rich feature representations from vast datasets, and can extract directly useful features for the prediction task.
Therefore, foundation models are a more convenient approach for the task than a training a ViT for the task to overcome the limitations of CNNs.
Transformer models are notorious for requiring extensive datasets for optimal performance.

{\setlength\tabcolsep{1 pt}
\begin{figure*}
\centering
\begin{tabular}{ccccccccc}
     \includegraphics[width=0.15\textwidth]{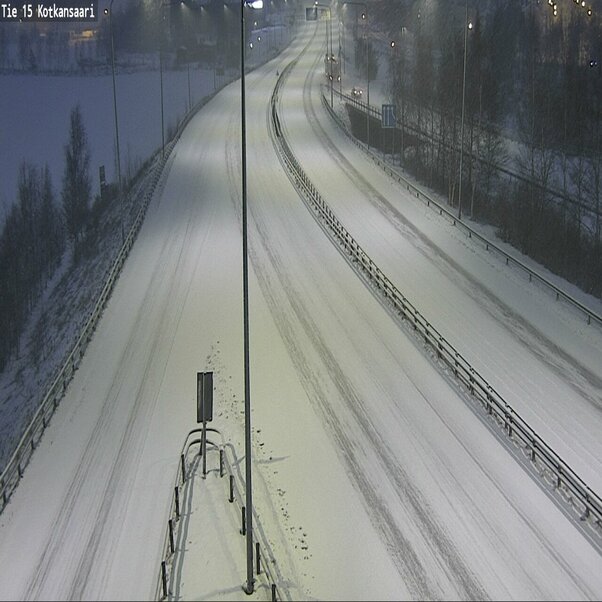} & 
      \includegraphics[width=0.15\textwidth]{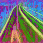} & 
       \hspace{0.01\textwidth} &
        \includegraphics[width=0.15\textwidth]{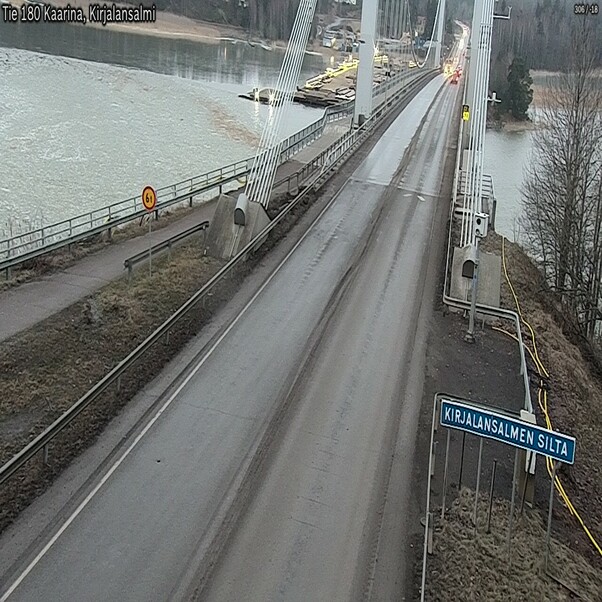} & 
         \includegraphics[width=0.15\textwidth]{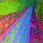} & 
          \hspace{0.01\textwidth} &
           \includegraphics[width=0.15\textwidth]{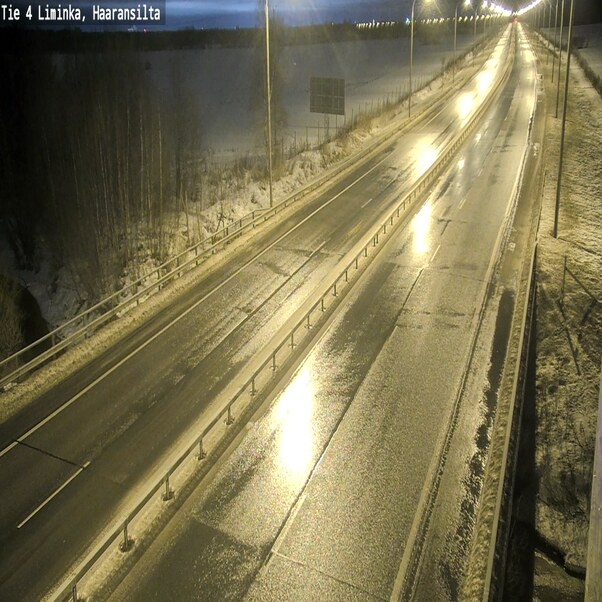} & 
            \includegraphics[width=0.15\textwidth]{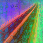} 
\end{tabular}
\caption{Image pairs demonstrating resized image samples and their respective DINOv2 patch tokens reduced to three feature dimensions with principal component analysis.}
\label{fig:dinov2_features}
\end{figure*}
}

\begin{figure*}
    \centering
    \includegraphics[width=0.95\textwidth]{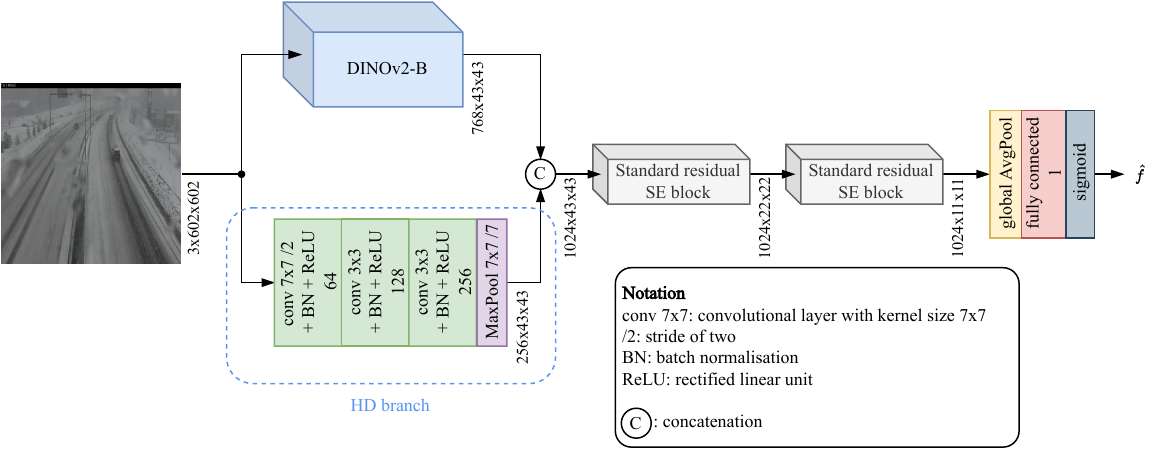}
    \caption{Proposed WCamNet architecture. Tensor sizes are reported for processing a single input image (batch size of one).}
    \label{fig:architecture}
\end{figure*}

The proposed WCamNet architecture is depicted in Figure \ref{fig:architecture}.
The moderately sized DINOv2 model, DINOv2-B, was chosen for the architecture.
During training of the network, the DINOv2 model remains frozen, while the weights of other layers are optimised.
When performing inference with the network, each RGB input image is resized to 602x602, and then fed to DINOv2 and a high-definition (HD) branch for parallel processing.
The HD branch consists of three CNN layers, which extract useful low-level features from the input image.
The final patch tokens produced by DINOv2 and the features from the HD branch are concatenated into a single tensor, which is fed through two consecutive standard residual squeeze-and-excitation (SE) blocks \cite{hu2018squeeze} with reduction ratios of 8.
These SE blocks enable weighing the relevant extracted features, along with providing additional convolutional feature extraction.
The output of the consecutive SE blocks is pooled with global average pooling, and fed through a fully connected layer with sigmoid activation.
This output is the friction factor prediction, $\hat{f}$, of the network. 

\subsection{Training and Testing}
For training and testing, the dataset was split with an approximate 50\%-15\%-35\% train, validation, and test split, respectively.
A high portion of testing data was chosen to ensure that the models were well generalisable to different camera views.
In the split, each roadside camera station was fully allocated to a specific split, ensuring that data from different cameras of a single station was not present in different splits.

To benchmark the performance of WCamNet, several other commonly applied deep learning models were trained and tested with the data as well.
ResNet50, ResNet152, VGG19, DINOv2-B with a linear head, as well as ViT-B/14 with DINOv2 pretraining were included to the benchmark for comparison.
The DINOv2-B with a linear head featured a fully connected layer which produced the prediction based on all the patch tokens as well as the final class token.
The linear head was trained on the training split, while the DINOv2 remained frozen.
VIT-B/14 with DINOv2 pretraining was essentially the DINOv2 model which was fully trained on the training split, using the DINOv2 weights as the starting point.
All models utilised a sigmoid activation at the end of the architecture for producing the final friction factor prediction.

WCamNet was trained for 15 epochs, and a Cosine Annealing learning rate scheduler with a warm restart every five epochs.
The CNN models, ResNet50, ResNet152 and VGG19, were trained for 30 epochs with a step-based learning rate scheduler reducing the learning rate to one tenth every ten epochs. 
The ViT-B/14 model was trained in an identical fashion.
DINOv2 with a linear head was trained for six epochs with the step-based learning rate scheduler decreasing the learning rate every two epochs.
All models were evaluated with PyTorch \cite{paszke2019pytorch} implementations, trained with mean squared error loss function, applying stochastic gradient descent with a momentum of 0.9, and a batch size of 16.
The initial learning rate and weight decay of each model were optimised via grid search.
These hyperparameters were selected based on performance on the validation split.

Images fed to the networks were resized to 602x602 pixels, and normalised based on the training set statistics.
During training, image augmentation was applied via random horizontal flipping, color jitter with a scale of 0.05, random rotations of -45 to 45 degrees, as well as random cropping with a padding of 64 pixels.

\section{Results}
\subsection{Accuracy Results}
In order to test the performance of the proposed WCamNet architecture, the accuracy of the model was evaluated on the testing split of the utilised dataset.
Several other models commonly utilised for the road surface condition prediction task were also evaluated for comparison.
Resulting accuracies of the models are presented in Table \ref{tab:accuracy} with the mean absolute error (MAE) and root-mean-square error (RMSE) metrics.

\begin{table}
\caption{Accuracies of the models on the test set. Bolded values indicate the best result in each column.}
\label{tab:accuracy}
\centering
\begin{tabular}{p{0.5\columnwidth} p{0.15\columnwidth} p{0.15\columnwidth}}
\\
\noalign{\hrule height 1.2pt}
  Method & MAE & RMSE \\
  \hline
  WCamNet (ours) & \textbf{0.150} & \textbf{0.195} \\
  ResNet50 & 0.184 & 0.233 \\
  ResNet152 & 0.201 & 0.260 \\
  VGG19 & 0.166 & 0.210 \\
  DINOv2-B w/ linear head & 0.172 & 0.219 \\
  ViT-B/14 (DINOv2 pretrained) & 0.183 & 0.232 \\
\noalign{\hrule height 1.2pt}
\end{tabular}
\end{table}

The acquired results demonstrate that WCamNet outperformed the other models in the accuracy benchmark. 
VGG19 was able to reach the second lowest errors in the benchmark.
Compared to VGG19, WCamNet reached a 10\% lower MAE and a 7\% lower RMSE.

\subsection{Ablation Study}
To analyse the rationality of the design choices in the WCamNet architecture, as well as the contribution of different parts of the network to the overall performance, an ablation study was carried out.
The training procedure was repeated with different configurations of the WCamNet architecture, and the accuracy results were evaluated on the testing split.
The ablations included changing the DINOv2-B model to the larger DINOv2-L model, removing the SE blocks from the architecture, as well as removing the HD branch from the architecture.
Acquired results for these ablations are presented in Table \ref{tab:ablations}.

\begin{table}
\caption{Ablation study of the proposed WCamNet. Bolded values indicate the best result in each column.}
\label{tab:ablations}
\centering
\begin{tabular}{p{0.6\columnwidth} p{0.1\columnwidth} p{0.1\columnwidth}}
\\
\noalign{\hrule height 1.2pt}
  Method & MAE & RMSE \\
  \hline
  WCamNet w/ DINOv2-L backbone & \textbf{0.155} & \textbf{0.197} \\
  WCamNet w/o SE blocks & 0.167 & 0.213 \\
  WCamNet w/o HD branch & 0.170 & 0.217 \\
\noalign{\hrule height 1.2pt}
\end{tabular}
\end{table}

The results demonstrate that the design choices behind WCamNet are reasonable and improve performance.
Accuracy with the larger DINOv2-L model in the architecture was highly similar as with the original architecture, yet the larger model results in increased computational cost.
Removal of SE blocks or the HD branch resulted in notable increase in errors, indicating that these elements are crucial for the performance of the model.

\section{Discussion}
The presented WCamNet model was shown to improve the state of the art of camera-based road surface friction estimation in winter conditions.
The hybrid architecture combining general visual features from DINOv2 and a CNN processing pipeline was shown to affect performance favourably.
Results demonstrated that the model achieved higher accuracy than models generally applied for the prediction task.
The performance gains demonstrated by the WCamNet model enable more reliable future utilisation of roadside camera-based road surface friction monitoring applications in intelligent transportation systems.

Although WCamNet was the most accurate computer vision model in the benchmark, the error values were still quite notable on an absolute scale.
This indicates that even computer vision cannot perfectly mimic the performance of the optical road friction sensor utilised as the ground truth.
Future research efforts are needed to more systematically map the reliability of computer vision-based road surface friction estimation, and its short-comings.

Another factor affecting the absolute error is the spatial ambiguity in the image data.
A single ground truth reading was used to label each image.
However, this is sub-optimal, since the road surface condition can vary within different parts of a single image.
For example, some sections of the road might be covered in snow, while other areas are not.
This causes ambiguity in the data, and has led to some authors proposing approaches with higher spatial resolution in the data \cite{roychowdhury2018machine, pesonen2023pixelwise}.
Nevertheless, this type of data is difficult to obtain, and therefore commonly a single label is used for the entire image.
Overall, the large quantity of the data used in this study should compensate for ambiguities in the data.

To further enhance camera-based road surface friction estimation in winter conditions, other weather data could be included as input in the prediction models.
Information such as temperature, precipitation, and humidity are generally available for most areas online, and these data could provide valuable contextual information for determining the road surface friction.
Fusion of camera images and weather information has been proposed in the past \cite{jonsson2011classification}, yet modern deep learning methods could likely demonstrate notable performance gains.
This type of data fusion could enable even higher accuracy in computer vision-based approaches, facilitating adoption of camera-based road monitoring methods in intelligent transportation applications.





\section*{ACKNOWLEDGMENT}

We acknowledge the funding provided by Henry Ford Foundation Finland, the computational resources provided by the Aalto Science-IT project, and the data provided by Fintraffic.


\bibliographystyle{ieeetr}
\bibliography{refs}

\end{document}